\documentclass{article}

\usepackage{microtype}
\usepackage{graphicx}
\usepackage{subcaption}
\usepackage{booktabs} 

\usepackage{amsmath,amsfonts,bm}

\def\eqref#1{equation~\ref{#1}}

\def\1{\bm{1}}

\DeclareMathAlphabet{\mathsfit}{\encodingdefault}{\sfdefault}{m}{sl}
\SetMathAlphabet{\mathsfit}{bold}{\encodingdefault}{\sfdefault}{bx}{n}

\usepackage[most]{tcolorbox}
\usepackage{enumitem}

\usepackage{hyperref}

\usepackage[preprint]{icml2026}

\usepackage{amsmath}
\usepackage{amssymb}
\usepackage{mathtools}
\usepackage{amsthm}

\usepackage[capitalize,noabbrev]{cleveref}

\theoremstyle{plain}
\newtheorem{theorem}{Theorem}[section]
\newtheorem{proposition}[theorem]{Proposition}
\newtheorem{lemma}[theorem]{Lemma}

\theoremstyle{definition}

\theoremstyle{remark}

\usepackage[textsize=tiny]{todonotes}

\usepackage{multirow}
\newcommand{\ie}{\mbox{\it{i.e.,\ }}}
\newcommand{\eg}{\mbox{\it{e.g.,\ }}}

\def\Snospace~{\S{}}

\newcommand{\sref}[2]{\hyperref[#2]{#1 \ref{#2}}}

 \DeclareMathAlphabet{\mathbbold}{U}{bbold}{m}{n}

\newcommand{\sys}{{\sc EntroPO}\xspace}
\setlist[itemize]{noitemsep, topsep=0pt}
\usepackage{tikz}
\usepackage{xcolor}
\usepackage{amsmath}
\usepackage{amssymb}
\usepackage{mathtools}
\usepackage{amsthm}
\usepackage{xspace}

\definecolor{yucky}{HTML}{a64d79}

\definecolor{darkgreen}{rgb}{0,0.40,0}
\definecolor{darkgrey}{gray}{0.5}
\definecolor{firebrick}{rgb}{0.698,0.133,0.133}
\usepackage[]{hyperref}
\hypersetup{                    
  colorlinks,
  linkcolor=firebrick,
  citecolor=darkgreen,
  urlcolor=firebrick
}

\theoremstyle{plain}

\usepackage{booktabs}
\usepackage{colortbl}
\definecolor{Orchid}{RGB}{218,112,214} %
\definecolor{purple}{RGB}{230,70,151}
\definecolor{lightgray}{gray}{0.9} %

\definecolor{bad}{RGB}{220, 20, 60} 
\definecolor{good}{RGB}{34, 139, 34}

\newcommand{\lite}{{\small\sc SWEBench-Lite}\xspace}
\newcommand{\verified}{{\small\sc SWEBench-Verified}\xspace}
\newcommand{\swebench}{{\small\sc SWEBench}\xspace}

\newif\ifrevision
\revisionfalse  

\newcommand{\revise}[1]{%
  \ifrevision
    {\color{red}#1}%
  \else
    #1%
  \fi
}

\icmltitlerunning{Building Coding Agents via Entropy-Enhanced Multi-Turn Preference Optimization}

\begin{document}

\twocolumn[
  \icmltitle{Building Coding Agents via Entropy-Enhanced \\
    Multi-Turn Preference Optimization}



  \icmlsetsymbol{equal}{*}

  \begin{icmlauthorlist}
    \icmlauthor{Jiahao Yu}{equal,nu}
    \icmlauthor{Zelei Cheng}{equal,c1,nu}
    \icmlauthor{Xian Wu}{meta}
    \icmlauthor{Xinyu Xing}{nu}
  \end{icmlauthorlist}

  \icmlaffiliation{nu}{Department of Computer Science, Northwestern University, Evanston, USA}
  \icmlaffiliation{c1}{Capital One, New York, USA}
  \icmlaffiliation{meta}{Meta AI, Bellevue, USA}

  \icmlcorrespondingauthor{Xian Wu}{xianwu123@meta.com}

  \icmlkeywords{Machine Learning, ICML}

  \vskip 0.3in
]



\printAffiliationsAndNotice{}  

\begin{abstract}

Test-time scaling (TTS) offers a promising path for solving complex software engineering tasks like \swebench, yet its efficacy is often bottlenecked by the \emph{diversity collapse} inherent in standard alignment methods like DPO. To address this, we introduce \sys, an entropy-enhanced preference optimization framework tailored for multi-turn, tool-using agents. We formulate the problem as an entropy-regularized MDP to derive the \sys-DPO and \sys-KTO objectives. Theoretically, we show that this approach promotes diversity specifically among \emph{correct} solutions, preventing the policy from converging on a narrow set of reasoning paths. To maximize TTS efficiency, we pair this training with a hybrid best-trajectory selection scheme. We validate \sys on \swebench leaderboard, where it achieves state-of-the-art results among open-weight models. Notably, our 30B model ranks 1st on \lite and 4th on \verified, suppressed only by models that are over 10$\times$ larger. These results highlight the importance of preserving diversity for effective test-time scaling and establish \sys as a robust method for building powerful, interactive coding agents.

\end{abstract}
\section{Introduction}\label{sec:intro}

Large Language Models (LLMs) have achieved impressive breadth across language understanding, coding assistance, and planning. Yet, they still struggle on complex, multi-step software engineering (SWE) tasks that demand reasoning over large codebases and coordinated tool use (\eg search, execution, and patching)~\citep{yang2024sweagent,wang2025openhands,agentless,antoniades2024swe,zhang2024autocoderover}. A promising line of work that improves performance is test-time scaling (TTS)---sampling more trajectories, searching deeper, and verifying candidates, which can uncover higher-quality solutions on challenging instances~\citep{snell2024scaling,beeching2024scalingtesttimecompute,yao2023tree,xu2024search}. However, TTS only helps if the model produces sufficiently \emph{diverse} candidates to explore meaningfully different solution modes.

Recent alignment methods, including Reinforcement Learning from Human Feedback (RLHF)~\citep{ouyang2022training,ziegler2019fine,bai2022training} and Direct Preference Optimization (DPO)~\citep{rafailov2023direct}, have been observed to inadvertently reduce generation diversity~\citep{kirk2023understanding,padmakumar2023does,kim2024knowledge,o2024attributing, murthy2025one}. This \emph{diversity collapse} limits the returns of TTS: when a model concentrates probability mass on a narrow set of responses for a given prompt, additional samples become redundant, and deeper search yields diminishing marginal gains. While recent work has attempted to adapt preference optimization to multi-turn tool trajectories (\eg M-DPO~\citep{xiong2025building}), these methods often fail to counteract the ``winner-take-all'' dynamics of preference learning, leading to collapsed policies that hinder the exploration required for complex problem solving. Prior efforts to preserve diversity typically target single-turn settings~\citep{slocum2025diverse,li2024preserving,wangbeyond,lanchantin2025diverse} or adjust decoding temperatures at inference time~\citep{renze2024effect}. These approaches do not directly address multi-turn, tool-using workflows, where diversity must be maintained \emph{throughout the trajectory} to encourage exploration of a sequence of tool calls and partial hypotheses.

To address this limitation, we introduce \sys, an entropy-enhanced preference optimization method for multi-turn SWE agents. 
Our approach augments the standard preference optimization objective with an entropy regularization term to preserve policy diversity and formulate it as an entropy-regularized Markov Decision Process framework.
Based on this framework, we formally derive the \sys-DPO and \sys-KTO loss functions in the multi-turn setting. Crucially, our theoretical analysis reveals that this entropy-regularized objective promotes diversity specifically among correct solutions: it assigns comparatively larger updates to high-utility trajectories that are underrepresented by the reference policy, effectively counteracting mode collapse. By optimizing over multi-turn interactions, \sys aligns the learning process with the sequential nature of coding tasks, teaching the model to build diverse and valid reasoning paths.

To maximize the performance gain of TTS, we pair \sys with a hybrid best-trajectory selection scheme. We combine (i) a learned verifier model that scores trajectories with (ii) model-free approaches that favor high-quality trajectories (\eg passing tests, trajectory steps). This hybrid selector improves sampling effectiveness and amplifies the gains from parallel rollouts.

We empirically validate \sys across a diverse suite of models from different families and sizes (up to 106B parameters) on \verified~\citep{chowdhury2024swebenchverified} and \lite~\citep{jimenezswe}. Our approach achieves state-of-the-art results among open-weight models, with our 30B model ranking 1st on \lite and 4th on \verified (surpassed only by models over 350B, which are 10x larger). Across all evaluations, \sys significantly outperforms standard DPO and KTO in the TTS setting. We also empirically demonstrate that our method effectively prevents the diversity collapse of the policy during preference learning, maintaining higher trajectory diversity compared to baselines. These results confirm that the entropy-preserving term is critical for maximizing the returns of test-time compute.

Our contributions are threefold:
\begin{itemize}[leftmargin=*]
    \item We propose \sys, an entropy-enhanced multi-turn preference optimization method that preserves policy diversity, and derive its corresponding DPO and KTO loss functions.
    \item We theoretically analyze our objective and show that our method promotes diversity over correct solutions, avoiding the convergence on a narrow set of ``winning'' trajectories common in standard preference learning.
    \item We present state-of-the-art results among open-weight models on \verified and \lite,  where TTS yields substantial gains through the diversity of our generated solutions.
\end{itemize}

By addressing the critical challenge of preserving diversity in multi-turn agents, our work paves the way for developing more powerful LLM-based tools capable of tackling real-world software engineering tasks.
\section{Related Work} \label{sec:related_work}

\textbf{LLM Post-training.} 
RLHF has become the standard approach for aligning LLMs with human preferences~\citep{ouyang2022training,ziegler2019fine,bai2022training,schulman2017proximal}, but the PPO-style online approach is computationally intensive, as it requires numerous interactions with a reward model or live environment to generate samples during training ~\citep{xu2024dpo, wei2025swerl}.
To reduce the compute cost, \emph{preference learning} methods such as DPO~\citep{rafailov2023direct} and KTO~\citep{ethayarajh2024kto}  replace explicit reward modeling and online RL with simpler, reward-free objectives that require far less compute.

However, most preference optimization methods remain limited to single-turn interactions. While recent extensions such as M-DPO \citep{xiong2025building} adapt these objectives for multi-turn tool trajectories, they primarily target mathematical reasoning. In long-context coding tasks, we observe that M-DPO suffers from \emph{diversity collapse}, hindering the exploration required for complex problem solving \citep{golubev2025training,gao2025beyond}. While recent methods~\citep{slocum2025diverse,wangbeyond,lanchantin2025diverse} have attempted to modify divergences, decouple KL components, or construct diversity-aware preference pairs to better control diversity, these approaches primarily focus on \emph{single-turn settings}. How to effectively extend these diversity-preserving mechanisms to complex, multi-turn tool-using workflows remains unexplored.

We address this gap with \sys, an entropy-enhanced framework that extends diversity-aware objectives to multi-turn preference optimization. Unlike prior approaches, \sys explicitly preserves exploration, which is crucial for effective test-time scaling on complex coding tasks. We also analyze the theoretical causes of DPO’s collapse and show how \sys mitigates these vulnerabilities.

\textbf{LLMs for Software Engineering.}
Repository-level SWE benchmarks such as the \swebench~\citep{jimenezswe,chowdhury2024swebenchverified} have accelerated progress on automated bug fixing and patch generation.
Agentic systems like SWE-agent~\citep{yang2024swe} introduced interfaces for repository navigation and code editing, while alternative pipelines (\eg Agentless)~\citep{agentless} achieved strong results with simpler localize-and-repair stages.
General-purpose agent frameworks such as OpenHands~\citep{wang2025openhands} provide open tooling for agents and show competitive performance on \swebench. These systems share core components (planning and tool-use) and must reason over large codebases via sequences of tool calls. This creates a critical need to maintain exploration and diversity throughout trajectories to enable more effective solution space exploration. Recent work have also explored agentic reinforcement learning (RL) to enhance complex reasoning ~\citep{deepswe2025, wei2025swerl}. These RL-based systems demonstrate significant potential for discovery but require substantial compute and complex online environment orchestration.

In contrast, our work applies trajectory-level preference optimization to align multi-turn tool-use behaviors. We choose offline preference optimization over online RLHF (\eg PPO~\citep{schulman2017proximal}, GRPO~\citep{shao2024deepseekmath}) primarily to reduce the compute and systems overhead of on-policy rollout generation and RL optimization. Furthermore, unlike standard preference optimization methods, which may suffer from diversity collapse, \sys includes an explicit entropy-preserving regularizer to prevent collapse over long-horizon sequences of tool calls.

\textbf{Test-Time Inference Strategies.} 
TTS strategies improve performance by sampling more candidates, searching deeper, and verifying outputs~\citep{snell2024scaling,beeching2024scalingtesttimecompute,yao2023tree,xu2024search}.
These include Best-of-$N$ sampling with verifier reranking and structured search methods such as Tree-of-Thoughts~\citep{yao2023tree}, which let models explore alternative reasoning paths and self-evaluate.
Recent approaches like R2E-gym~\citep{jain2025r2e} further enhance these methods by introducing hybrid selectors to improve verification robustness. However, these techniques primarily improve post-generation selection without addressing the underlying generation process. This limitation is critical because the effectiveness of TTS strategies relies on the diversity of candidate solutions~\citep{snell2025scaling,wang2025diversified,huang2025best}. When generations collapse to a narrow solution space, additional samples and deeper search provide diminishing gains. 

In this work, we combine the generation-time diversity enhancement from \sys with a hybrid best-trajectory selector. It leverages the increased diversity from \sys while maintaining robustness to verifier errors, yielding stronger empirical gains as test-time compute scales.

\section{Proposed Technique} \label{sec:method}

\begin{figure*}[t]
    \centering
    \includegraphics[width=0.9\textwidth]{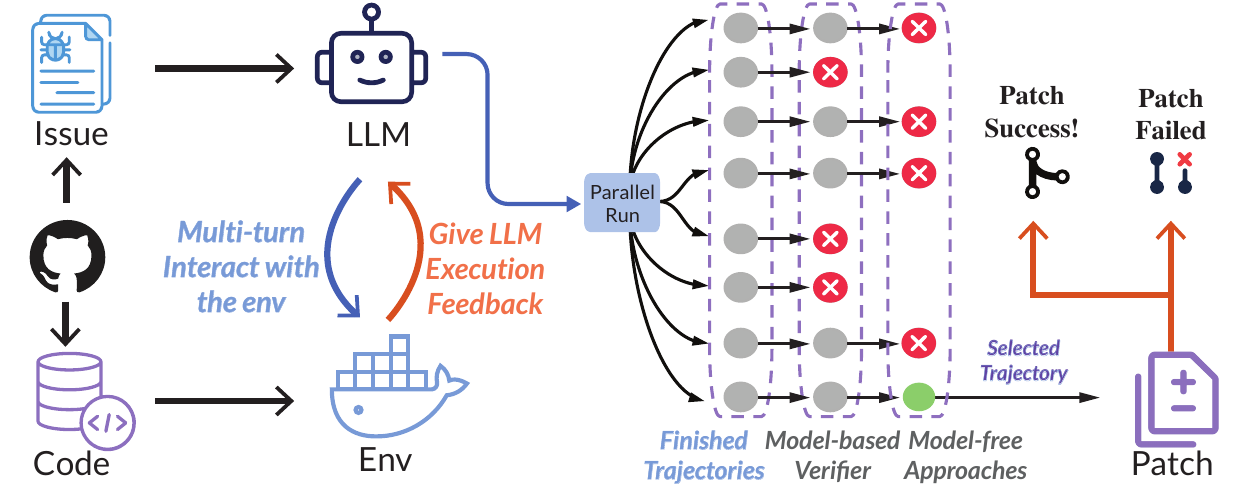}
    \caption{\textbf{Overview of \sys with TTS.} Given an issue and a repository, an LLM agent interacts with a sandboxed environment over multiple turns, receiving execution feedback. We run parallel rollouts to produce a pool of candidate trajectories. A hybrid selector ranks trajectories using a model-based verifier and model-free approaches, and selects the best trajectory to submit.}
    \label{fig:workflow}
\end{figure*}

We propose \sys, an entropy-enhanced preference optimization framework for multi-turn, tool-using coding agents. As shown in~\autoref{fig:workflow}, we use an agent that follows the standard SWE workflow to interact with a sandboxed repository environment and receive execution feedback at each turn. For TTS, we launch parallel rollouts to collect a set of trajectories for each issue. These trajectories are then scored by a hybrid selector that combines a model-based verifier with model-free approaches. The top-scoring trajectory is selected to submit as the final patch and is evaluated by the benchmark tests. Our core contributions are a novel \emph{training objective} that preserves trajectory-level diversity and a \emph{hybrid selection} mechanism that effectively exploits this diversity. To do so, we augment the standard preference optimization objective with an explicit entropy regularization term, which directly encourages the policy to maintain a broader distribution over potential solutions. For the agent itself, we build upon a standard scaffold, avoiding the introduction of new tool schemas.

\subsection{Problem Setup and Assumption} \label{sec:problem}

We frame the multi-turn coding task as a finite-horizon episodic Markov Decision Process (MDP), $\mathcal{M} = \langle \mathcal{S}, \mathcal{A}, H, \mathbb{P}, d_0, r \rangle$. Here, $\mathcal{S}$ is the state space, $\mathcal{A}$ is the action space, $H$ is the maximum number of turns (horizon), $\mathbb{P}$ is the transition dynamics, $d_0$ is the initial state distribution, and $r$ is a reward function.

An initial state $s_1 = x \sim d_0$ represents a coding problem statement. At each step $h \in \{1, \dots, H\}$, the agent's policy $\pi(a_h|s_h)$ observes the current state $s_h$, which contains the full interaction history, and generates an action $a_h$ (\eg a bash command). The environment executes $a_h$, returns an observation $o_h$ (\eg compiler output, test results, or tool feedback), and transitions to the next state $s_{h+1} = (s_h, a_h, o_h)$. This interaction results in a trajectory $\tau = (x, a_1, o_1, \dots, a_H)$. We define the interaction completion $y$ as the sequence of generated actions and observations, allowing us to denote the trajectory as the concatenation $\tau = [x, y]$. 

We assume access to a preference dataset $\mathcal{D}$ consisting of pairs of trajectories. 
Preferences are provided at the \textbf{trajectory level}\footnote{By construction, the reward can equivalently be written as $r(x, y) = r(s_H, a_H)$.} 
by an automated oracle that evaluates whether the final code produced by a trajectory passes a suite of unit tests; 
trajectories that pass are strictly preferred over those that fail. 
Following prior work \citep{ouyang2022training, xiong2025building}, we model preference probabilities using the Bradley--Terry model \citep{bradley1952rank}:
\begin{equation}
P(y^+ \succ y^- | x, y^+, y^-) = \sigma(r(x, y^+) - r(x, y^-))
\end{equation}
where $r$ is a latent reward function and $\sigma(\cdot)$ denotes the sigmoid function.

\subsection{Algorithmic Formulation of \sys} \label{sec:algo}

Standard preference optimization methods, such as DPO~\citep{rafailov2023direct}, induce \emph{diversity collapse}~\citep{murthy2025one}, where the policy converges to a narrow set of solutions. This is particularly detrimental for reasoning tasks where diverse exploration is required for effective TTS.

To counteract this, we augment the standard preference optimization objective with a weighted entropy regularization term, $\lambda H(\pi)$. This term directly penalizes low-entropy policies, encouraging the model to maintain a broader distribution over viable action sequences. This ensures the model not only learns what constitutes a high-quality trajectory but also retains the stochasticity needed to explore a diverse set of candidates at inference time, thereby maximizing the benefit of TTS.

Our full objective, framed as a regularized MDP, is to find the optimal policy $\pi^*$  that maximizes the expected reward while encouraging diversity and staying close to a reference policy $\pi_{\text{ref}}$:
\begin{equation}
\label{eqn:obj}
\begin{split}
&\max_{\pi} \mathbb{E}_{x \sim d_0, a_{h} \sim \pi(\cdot|s_{h}), o_{h} \sim \mathbb{P}_h(\cdot|s_{h}, a_{h})} \\
&\Big[ r(x,y) - \gamma \cdot D_{KL}(\pi || \pi_{\text{ref}}) + \lambda \cdot H(\pi(\cdot|x)) ) \Big],
\end{split}
\end{equation}
where the parameter $\lambda$ promotes diversity, and the coefficient $\gamma$ penalizes the deviation between the learned policy and the referenced policy $\pi_{\text{ref}}$. $H(\pi(\cdot|x)) = -\pi(\cdot|x)^T \log \pi(\cdot|x)$ denotes the entropy of the learned policy $\pi$. 

To optimize the \sref{Eqn.}{eqn:obj},  we first define the regularized value function for policy $\pi$ as:
\begin{equation}
\label{eqn:value_function}
\small 
V^{\pi}_{h}(s) = \mathbb{E}_{\pi} \Bigg[ \sum_{t=h}^{H} \bigg( r(s_h, a_h) \\
- \zeta \log \frac{\pi(a_h|s_h)}{\pi_{\text{ref}}(a_h|s_h)^{\frac{\gamma}{\zeta}}} \bigg) \Bigg| s_h = s \Bigg],
\end{equation}
where $\zeta = \lambda + \gamma$. Note that \sref{Eqn.}{eqn:value_function} can be derived by expanding the KL divergence and entropy terms in \sref{Eqn.}{eqn:obj} into their logarithmic forms and rearranging the corresponding items. By definition, we set $V_{H+1}(s) = 0$.

To leverage standard results from the entropy-regularized MDP, we define an augmented reward function $r'(s,a) = r(s,a) + \gamma\log \pi_{\text{ref}}(a|s)$. This formulation absorbs the reference policy constraint, allowing the regularized $Q$-function to be written as:
\begin{equation}
Q^{\pi}_{h}(s,a) = r'(s,a) + \mathbb{E}_{s' \sim p(\cdot|s,a)} [V^{\pi}_{h+1}(s')].
\end{equation}
The relationship between the value function and the $Q$-function follows the standard consistency equation:
\begin{equation}
V^{\pi}_{h}(s) = \mathbb{E}_{a \sim \pi(\cdot|s)} [ -\zeta \log \pi(a|s) + Q^{\pi}_{h}(s,a) ].
\end{equation}
Under this formulation, both the value function and the $Q$-function are consistent with the maximum entropy policy optimization framework~\citep{ziebart2010modeling}. Consequently, the regularized optimal policy $\pi^*$, which maximizes $V^{\pi}_{h}(s)$, can be expressed in terms of the optimal $Q$-function $Q^*_{h}$ and value function $V^*_{h}$ as follows:
$\pi^*(a|s) = \exp\left\{ (Q^*_{h}(s,a) - V^*_{h}(s)) / \zeta \right\}.$ To derive the preference optimization objective, we establish a connection between the rewards and the optimal policy, inspired by DPO. By rearranging the optimality condition  above and summing over $h \in [H]$, we obtain the following lemma:
\begin{lemma}[Reward Sum Decomposition]
\label{lemma:reward}
Consider a deterministic MDP, the total accumulated reward can be expressed in terms of the optimal policy $\pi^*$, the reference policy $\pi_{\text{ref}}$, and the initial value function $V^*_{1}(s_1)$ as follows:
\begin{equation}
\sum_{h=1}^{H} r(s_h, a_h) = \sum_{h=1}^{H} \zeta \log \frac{\pi^*(a_h|s_h)}{\pi_{\text{ref}}(a_h|s_h)^{\frac{\gamma}{\zeta}}} + V^*_{1}(s_1).
\end{equation}
\end{lemma}

The proof of ~\sref{Lemma}{lemma:reward} is in \autoref{app:proof_lemma1}. Combining this result with the Bradley-Terry model allows us to derive the optimal policy and loss function for the multi-turn objective, as summarized in the following theorem.

\begin{theorem}
\label{theorem:two_turn}
Consider the multi-turn setting, assuming the preferences follow a Bradley-Terry model and the policy optimization objective is given in \sref{Eqn.}{eqn:obj}, we have the following DPO-style loss function:
\begin{multline}
L_{\sys-DPO}(\theta) = \\
    - \mathbb{E}_{(x, \tau^+, \tau^-) \in \mathcal{D}} \Bigg[ \log \sigma \Bigg( \zeta \sum_{h=1}^{H} \bigg( \log \frac{\pi_{\theta}(a_h^+|s_h^+)}{\pi_{\text{ref}}(a_h^+|s_h^+)^{\frac{\gamma}{\zeta}}}\\
    - \log \frac{\pi_{\theta}(a_h^-|s_h^-)}{\pi_{\text{ref}}(a_h^-|s_h^-)^{\frac{\gamma}{\zeta}}} \bigg) \Bigg) \Bigg],    
\end{multline}
where $(\tau^+, \tau^-)$ represents a preference pair.
\end{theorem}

Theoretically, under the Bradley-Terry assumption, optimizing this \sys-DPO loss yields the same optimal policy as maximizing the ground-truth reward in \sref{Eqn.}{eqn:obj}, provided the implicit reward is perfectly recovered. We also derive our KTO-style loss variant, \sys-KTO, in \autoref{app:kto_loss}.

\subsection{Diversity Analysis of \sys}
\label{sec:div_analysis}
In this subsection, we theoretically analyze the diversity-promoting properties of \sys. First, we observe that the completion $y$ consists of observation $o$ from the environment and the model's response $y'$. Since the code execution environment is deterministic,  the probability of a completion $P(y|x)$ is equivalent to $\pi(y'|x)$ (see Appendix~\ref{app:proof_bandit}). Consequently, we parameterize $P(y|x)$ directly via a generative model $\tilde{\pi}(y|x)$. This reduction allows us to derive the optimal solution for $\tilde{\pi}(y|x)$ by maximizing the following objective:
\begin{lemma}
\label{lemma:optimal_policy}
With our re-parameterization, the objective in \sref{Eqn.}{eqn:obj} can be transformed to 
\begin{equation}
\begin{split}
&\max_{\tilde{\pi}} \mathbb{E}_{x \sim d_0, y \sim \tilde{\pi}(\cdot|x)} \\
&[r(x,y) + \zeta \cdot H(\tilde{\pi}(\cdot|x)) - \gamma \cdot H(\tilde{\pi}, \tilde{\pi}_{\mathrm{ref}})],  
\end{split}
\label{eqn:obj_reparam}
\end{equation}
where $H(\tilde{\pi}, \tilde{\pi}_{\mathrm{ref}}) = -\tilde{\pi}(\cdot|x)^T \log \tilde{\pi}_{\mathrm{ref}}(\cdot|x)$ represents the cross entropy between the learned policy $\tilde{\pi}$ and the referenced policy $\tilde{\pi}_{\mathrm{ref}}$. The optimal policy is then given by 
\begin{equation}
\tilde{\pi}^*(y|x) \propto  \tilde{\pi}_{\mathrm{ref}}(y|x)^{\frac{\gamma}{\zeta}} \exp\left(\frac{r(x,y)}{\zeta}\right). 
\end{equation}
\end{lemma}

The derivation of the closed-form optimal solution $\tilde{\pi}^*$ can be found in Appendix~\ref{app:proof_optimal} \footnote{While treated $\tilde{\pi}$ as unconstrained, \sref{Eqn.}{eqn:obj_reparam} is technically a constrained optimization over $\tilde{\pi}$; details in Appendix~\ref{app:proof_optimal}.}. By analyzing the relative increase in log-probability from the optimal policy to the reference policy, we show why the standard DPO method can lead to \emph{diversity collapse} in our multi-turn setting, whereas \sys counteracts this effect by preferentially upweighting correct responses that are underrepresented under the reference policy.

\begin{proposition}[Mitigation of Diversity Collapse via Inverse Probability Weighting]
\label{proposition:diversity}
Consider two completions $y_1, y_2$ yielding identical positive rewards $r(x, y_1) = r(x, y_2)=1$.
Standard DPO ($\gamma = \zeta$)~\citep{rafailov2023direct,xiong2025building} applies a uniform scaling factor to both, preserving the relative likelihood ratio of the reference policy ($\frac{\tilde{\pi}^*(y_1)}{\tilde{\pi}^*(y_2)} = \frac{\tilde{\pi}_{\mathrm{ref}}(y_1)}{\tilde{\pi}_{\mathrm{ref}}(y_2)}$) and failing to correct existing diversity collapse.
In contrast, \sys ($\gamma < \zeta$) induces a relative log-likelihood gain strictly inversely proportional to the reference likelihood.
\end{proposition}

We provide a proof in Appendix~\ref{app:proof_claim}. Consequently, \sys counteracts diversity collapse by assigning strictly greater probability amplification to correct responses that are underrepresented in $\tilde{\pi}_{\mathrm{ref}}$ relative to those that are common. 
\subsection{Training and Inference Pipeline} \label{sec:pipeline}

Our approach integrates a two-stage training pipeline with a robust test-time scaling strategy. We first teach the base model to use tools reliably via SFT on teacher-generated trajectories, then apply preference learning with our entropy-enhanced objective (detailed training procedures and data generation steps are provided in \autoref{app:training_inference_details}). At inference, we scale test-time compute by generating $N$ parallel rollouts and selecting the best solution using a hybrid selector. This selector applies four filters: validity checks, regression testing, model-based verification, and length-based preference (details of the hybrid selector are provided in \autoref{app:training_inference_details}). This pipeline effectively prunes low-quality solutions and amplifies the gains from parallel scaling.

\section{Experiments} \label{sec:exp}
In this section, we show the performance of \sys on benchmarks with R2E~\citep{jain2025r2e} agent scaffold. \autoref{sec:exp_setup} details datasets, models, and training/inference configurations, including verifier training and the TTS budget. \autoref{sec:exp_results} presents the main results and comparisons to official leaderboard submissions. \autoref{sec:exp_ablation} analyzes scaling with the number of parallel rollouts $N$ and ablates \sys components and hyperparameters. Additional implementation details are provided in \autoref{app:implementation}. Finally, \autoref{app:additional_experiments} provides further analysis on sampling temperature sensitivity, the impact of entropy regularization on \sys-KTO, and ablations of step heuristics for the hybrid selector.

\subsection{Implementation Details} \label{sec:exp_setup}
\textbf{Datasets.} For SFT tuning, we use the SWE-Smith dataset~\citep{yang2025swe}, which does not rely on oracles. For preference learning and verifier model training, we use the R2E-Gym-subset~\citep{jain2025r2e}, whose oracles provide trajectory-level utilities. We evaluate on \verified and \lite, reporting resolve rates based on their official protocols. All performance numbers are pass@$1$ and no hint or web search is used.

\textbf{Models.} We evaluate a diverse set of models from three different families, with sizes ranging from 4B to 106B: Qwen3-4B-Instruct-2507~\citep{qwen3technicalreport}, Gemma-3-27b-it~\citep{team2025gemma}, Qwen3-Coder-30B-A3B-Instruct, and GLM-4.5-Air-106B~\citep{zeng2025glm}. The verifier is trained with Qwen3-Coder-30B-A3B-Instruct for its strong coding quality and high token throughput.

\textbf{Training and Inference.} We train with LLaMAFactory~\citep{zheng2024llamafactory} and set the maximum training sequence length to 18{,}000 tokens to accommodate long SWE trajectories.
To manage memory, we use QLoRA~\citep{dettmers2023qlora} for GLM-4.5-Air and LoRA~\citep{gao2021lora} for the other models.
Unless noted, \sys uses $\zeta=0.11, \gamma=0.1$ and the sensitivity to $\zeta / \gamma$ is reported in \autoref{sec:exp_ablation}.
During SFT and preference training, we mask system and user prompts and make the LLM response as the learning target.
At inference, we allow up to 200 environment interactions per rollout and a maximum sequence length of 131{,}072 tokens, with temperature $0.7$ and \texttt{top\_k} $=20$.
For test-time scaling, we run $N=16$ parallel rollouts for open-weight models.
To account for sampling randomness, all experiments on open-weight models are run three times, and we report the mean $\pm$ standard deviation.

\subsection{Main Results} \label{sec:exp_results}

\begin{table*}[t]
\centering
\small
\caption{\textbf{Resolve rate (pass@1, \%) on benchmarks.} Results are mean $\pm$ std over three runs and the TTS uses $N=16$ parallel rollouts. For each model, the best result is highlighted in gray.}
\label{tab:results_internal}
\resizebox{2.0\columnwidth}{!}{
\begin{tabular}{l|cccccc}
\toprule
Model & Origin & SFT & \sys-KTO & \sys-DPO & \sys-KTO+TTS & \sys-DPO+TTS \\
\midrule
\multicolumn{7}{c}{\textit{\verified}} \\
\midrule
\rowcolor{gray!8} Qwen3-4B & 1.7 ($\pm$ 0.1) & 2.4 ($\pm$ 0.3) & 5.2 ($\pm$ 0.4) & 4.9 ($\pm$ 0.7) &\cellcolor{darkgrey!30} 11.5 ($\pm$ 0.6) & 11.1 ($\pm$ 0.3) \\
Gemma-3-27b & 7.1 ($\pm$ 0.5) & 7.0 ($\pm$ 0.4) & 10.1 ($\pm$ 0.4) & 10.5 ($\pm$ 0.3) & 17.6 ($\pm$ 0.3) &\cellcolor{darkgrey!30} 17.7 ($\pm$ 0.4) \\
\rowcolor{gray!8} Qwen3-Coder-30B & 37.7 ($\pm$ 0.2) & 43.8 ($\pm$ 0.8) & 51.6 ($\pm$ 0.7) & 49.8 ($\pm$ 0.4) &\cellcolor{darkgrey!30} 59.4 ($\pm$ 0.3) & 57.7 ($\pm$ 0.7) \\
GLM-4.5-Air & 51.4 ($\pm$ 0.2) & 51.5 ($\pm$ 0.8) & 53.5 ($\pm$ 0.7) & 52.5 ($\pm$ 0.7) &\cellcolor{darkgrey!30} 58.7 ($\pm$ 0.1) & 57.5 ($\pm$ 0.4) \\
\midrule
\multicolumn{7}{c}{\textit{\lite}} \\
\midrule
\rowcolor{gray!8} Qwen3-4B & 1.2 ($\pm$ 0.3) & 1.3 ($\pm$ 0.5) & 4.8 ($\pm$ 0.4) & 4.7 ($\pm$ 0.5) & 10.0 ($\pm$ 0.8) &\cellcolor{darkgrey!30} 10.4 ($\pm$ 0.4) \\
Gemma-3-27b & 6.0 ($\pm$ 0.8) & 5.9 ($\pm$ 0.4) & 10.4 ($\pm$ 0.2) & 10.4 ($\pm$ 0.7) &\cellcolor{darkgrey!30}  14.6 ($\pm$ 0.6) & 14.4 ($\pm$ 0.7) \\
\rowcolor{gray!8} Qwen3-Coder-30B & 28 ($\pm$ 0.3) & 33.9 ($\pm$ 0.3) & 44.0 ($\pm$ 0.3) & 43.7 ($\pm$ 0.8) &\cellcolor{darkgrey!30} 49.2 ($\pm$ 0.7) & 48.2 ($\pm$ 0.7) \\
GLM-4.5-Air & 43.5 ($\pm$ 0.6) & 43.9 ($\pm$ 0.4) & 44.9 ($\pm$ 0.4) & 44.6 ($\pm$ 0.5) & \cellcolor{darkgrey!30} 48.4 ($\pm$ 0.1)  & 47.9 ($\pm$ 0.3)\\
\bottomrule
\end{tabular}
}
\end{table*}

\textbf{Comparison with Baselines.} We compare \sys against (i) the base model, (ii) the SFT-tuned model, and (iii) representative preference-optimization baselines (\ie M-DPO/KTO~\citep{xiong2025building}) under the same scaffolds and budgets. We compare \sys with the base model and the SFT-turned model in \autoref{tab:results_internal}. A detailed comparison with M-DPO and M-KTO, demonstrating the advantage of our entropy-regularized objective for test-time scaling, is presented in \autoref{fig:tts_curve} and \autoref{fig:tts_curve_dpo}.

As shown in \autoref{tab:results_internal}, \sys consistently outperforms both the original and SFT-tuned models across all benchmarks, even without TTS. For models like Gemma-3-27b and GLM-4.5-Air, standard SFT yields minimal gains over the base models. In contrast, \sys delivers substantial improvements, which we attribute to its entropy-regularized objective that preserves policy diversity. This increased diversity is critical for effective exploration and better generalization to unseen problems. When combined with TTS, the performance gains are further amplified, aligning with our theoretical analysis that diversity is key to maximizing the benefits of test-time compute.
\revise{
    Note that here the results for Qwen3-Coder-30B are different from the original paper because we use the R2E scaffold for consistency instead of the OpenHands scaffold and a much shorter maximum context length due to inference cost consideration.
}

The impact of \sys is particularly evident for smaller models. For instance, the Qwen3-4B model's performance is negligible after SFT (1.7\% on \verified and 1.2\% on \lite), indicating a failure to learn the task. However, with \sys and TTS, its resolve rate surpasses 10\%—a remarkable improvement that demonstrates the potential of our approach to make smaller, more efficient models viable for complex SWE tasks.

For larger models, \sys also achieves significant gains. The Qwen3-Coder-30B model trained with \sys-KTO+TTS reaches 59.4\% on \verified and 49.2\% on \lite, establishing a strong performance baseline. We note that for GLM-4.5-Air, the improvements from \sys are less pronounced compared to Qwen3-Coder-30B. This is likely due to the use of QLoRA for fine-tuning GLM-4.5-Air, which is not as effective as LoRA. 

\begin{table*}[t]
    \centering
    \small
    \caption{\textbf{Resolve rate (pass@1, \%) compared to representative \swebench leaderboard submissions.} Our entries use the R2E scaffold, and we report the best single run for comparability. Budgets and scaffolds may vary across submissions. }
    \label{tab:leaderboard_comparison}
    \resizebox{2.0\columnwidth}{!}{
    \begin{tabular}{ll|ccc}
        \toprule
        Submission & Model & Model Size & \verified $\downarrow$ & \lite $\downarrow$  \\
        \midrule
        \multicolumn{5}{c}{\textit{Closed Weight Models}} \\
        \midrule
        Refact.ai & Claude3.7/o3/o4-mini & - & 74.4 & 60.0 \\
        SWE-agent & Claude 4 Sonnet & - & 66.6 & 56.7 \\
        SWE-agent & Claude 3.7 Sonnet & - & 62.4 & 48.0 \\
        \midrule
        \multicolumn{5}{c}{\textit{Open Weight Models}} \\
        \midrule
        OpenHands & Qwen3-Coder & 480B & 69.6 & - \\
        OpenHands & Kimi K2 & 1T & 65.4 & - \\
        OpenHands & GLM-4.5 & 355B & 64.2 & - \\
        DeepSWE-TTS & Qwen3 & 32B & 58.8 & - \\
        Skywork-SWE-TTS & Qwen2.5 & 32B & 47 & - \\
        CodeFuse-CGM & Qwen2.5 & 72B & - & 44.0 \\
        KGCompass & DeepSeek V3 & 671B & - & 36.7 \\
        SWE-fixer & Qwen2.5 & 72B & 24.7 & 32.8 \\
        Moatless & Deepseek V3 & 671B & - & 30.7 \\
        \midrule
        \multicolumn{5}{c}{\textit{Our Methods}} \\
        \midrule
        \rowcolor{gray!8} \sys-KTO-TTS & Qwen3-Coder & 30B & 60.4 & 49.7 \\
        \rowcolor{gray!8} \sys-KTO & Qwen3-Coder & 30B & 52.2 & 45.0 \\
        \bottomrule
    \end{tabular}
    }
\end{table*}

\textbf{Comparison to the Official \swebench Leaderboard.}
In \autoref{tab:leaderboard_comparison}, we compare our best-performing model against submissions on the official \swebench leaderboard. Our results are highly competitive, particularly among open-weight models. On \verified, our 30B parameter \sys-KTO-TTS model achieves a 60.4\% resolve rate, surpassed only by models with over 10x more parameters (\eg $>$350B). On \lite, the same model sets a new state-of-the-art for open-weight models at 49.7\%, with our non-TTS version securing the second-highest rank.

Crucially, \sys-KTO-TTS outperforms other TTS-based submissions like DeepSWE-TTS~\citep{deepswe2025} and Skywork-SWE-TTS~\citep{zeng2025skywork}. This highlights the effectiveness of our entropy-preserving training, which preserves the policy diversity essential for maximizing TTS gains. Note that DeepSWE-TTS is online RL-based, showing that our entropy-preserving offline preference learning can be more effective than online RL.
Compared to closed-weight models, our results are competitive with top-tier models like Claude 3.7 Sonnet, demonstrating that \sys can significantly narrow the performance gap with commercial models.

\subsection{Ablation Studies} \label{sec:exp_ablation}
\revise{
In this section, we perform ablation studies to investigate the impact of different components of \sys and the sensitivity of the hyperparameters. We mainly focus on the \sys-KTO, as KTO requires fewer GPU memory compared with DPO, as DPO takes a pair of trajectories to calculate the gradient during training.
}

\begin{figure*}[t]
    \centering
    \includegraphics[width=0.48\textwidth]{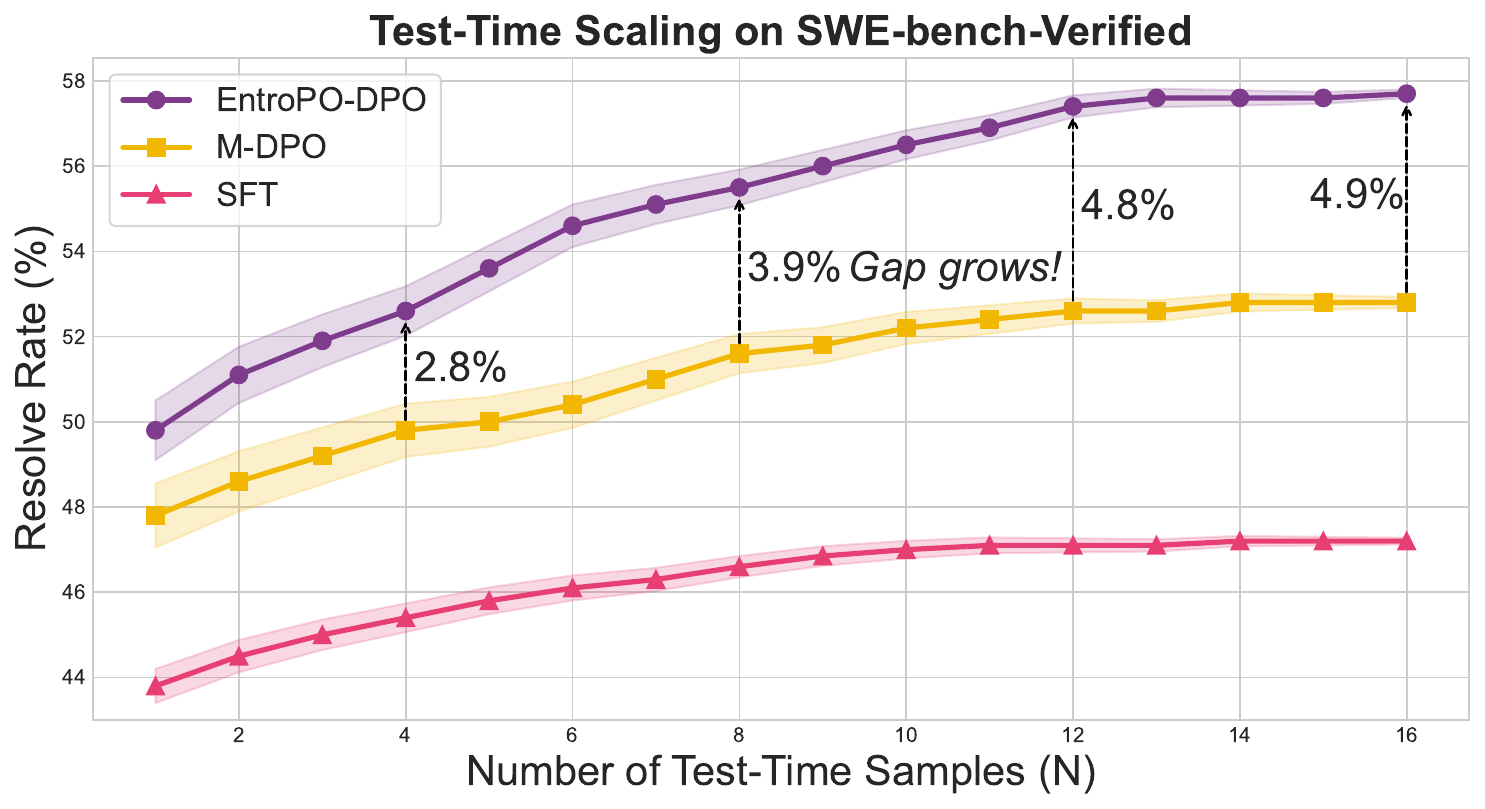}
  \includegraphics[width=0.48\textwidth]{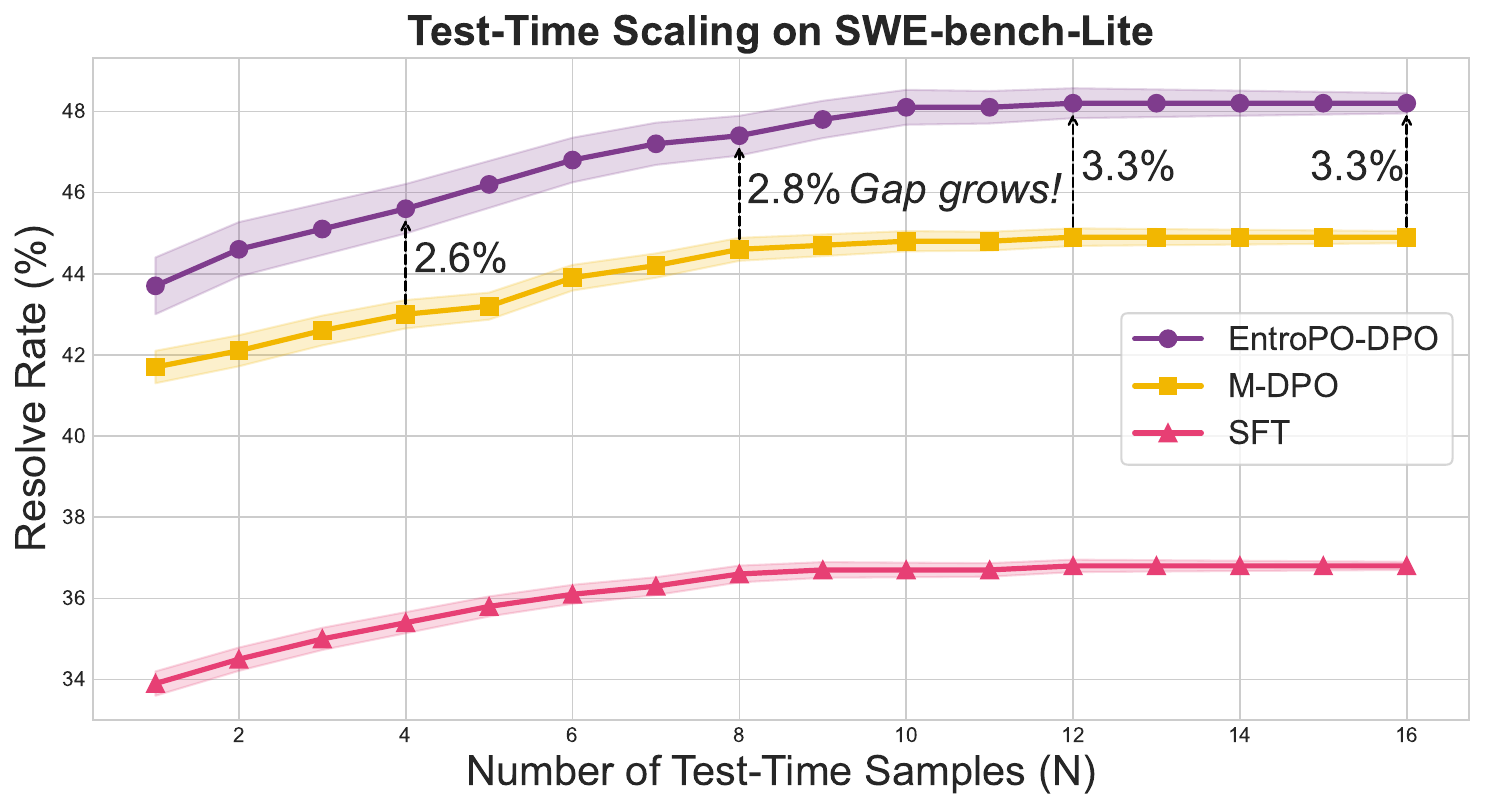}
    \caption{\textbf{The Impact of Entropy Regularization on Test-Time Scaling.} Performance of \sys-DPO, M-DPO, and SFT on \verified (left) and \lite (right) as the number of parallel rollouts ($N$) increases. \sys's entropy regularization consistently yields better scaling.}
    \label{fig:tts_curve}
\end{figure*}

\begin{figure*}[t]
    \centering
    \includegraphics[width=1.0\textwidth]{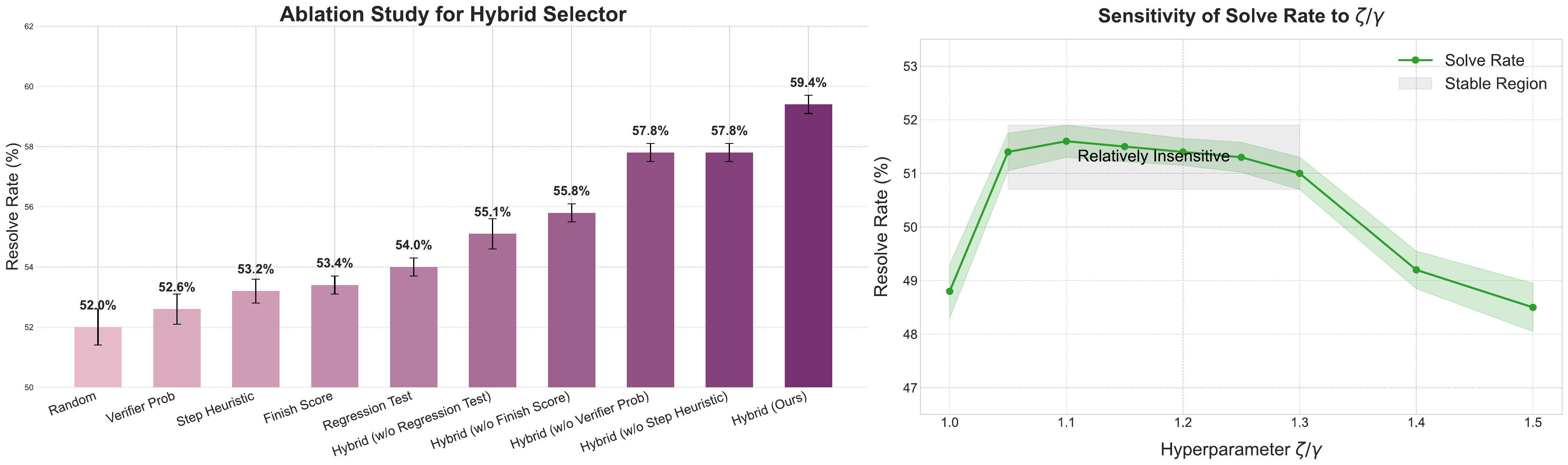}
    \caption{\textbf{Ablation Studies on \verified.} (Left) Performance contribution of each component in our hybrid selector at $N=16$. (Right) Sensitivity analysis of the hyperparameter $\zeta / \gamma$ for \sys-KTO. For visualization convenience, we plot the inverse of $\gamma / \zeta$ from \autoref{theorem:two_turn}. }
    \label{fig:ablation_study}
\end{figure*}

\textbf{Impact of Entropy Regularization.} To isolate the benefit of our entropy-preserving objective, we compare \sys-DPO against the M-DPO~\citep{xiong2025building} and SFT on the Qwen3-Coder-30B model. 
As shown in \autoref{fig:tts_curve}, \sys-DPO consistently outperforms both baselines. 
\revise{
  When $N=1$, \sys-DPO can outperform both SFT and M-DPO, showing its advantage even \emph{without TTS}.
}
The performance gap between \sys-DPO and M-KTO widens with larger $N$, confirming that explicit diversity preservation is critical for maximizing the gains from TTS. Both preference-based methods outperform SFT, which aligns with findings that SFT can harm generalization~\citep{chu2025sft}.
\revise{
    We conduct a similar experiment on the \sys-KTO model and the M-KTO model, and the results are shown in \autoref{fig:tts_curve_dpo}. The results show a similar trend to the \sys-DPO experiment.
}

\textbf{Hybrid Selector Components.} We analyze the contribution of each component of our hybrid selector at $N=16$. The left plot in \autoref{fig:ablation_study} shows that removing any single component degrades performance, while the full hybrid selector achieves the best results. This confirms that combining a learned verifier with model-free approaches is the most effective strategy.

\textbf{Sensitivity to $\zeta / \gamma$.} The right plot in \autoref{fig:ablation_study} shows the performance of \sys-KTO across different values of the hyperparameter $\zeta / \gamma$. The model is robust to a reasonable range of $\zeta / \gamma$, indicating that it is not a sensitive hyperparameter. Performance degrades only when $\zeta / \gamma$ becomes excessively large, causing the training gradients to vanish.

\textbf{Impact of Temperature on Performance.} We also investigate whether increasing sampling temperature can replicate the benefits of \sys. As detailed in \autoref{app:additional_experiments}, simply raising the temperature fails to deliver comparable performance gains. Instead, it degrades performance by introducing excessive sampling randomness, confirming that temperature tuning is no substitute for the principled entropy regularization of \sys.

\textbf{Diversity Analysis.} We analyze the diversity of generated trajectories by \sys-KTO and \sys-DPO compared with the M-DPO and M-KTO models. The detailed analysis is shown in \autoref{app:diversity_analysis}. These empirical results align with our theoretical analysis in \autoref{sec:div_analysis}, showing that our entropy-regularized objective promotes higher diversity compared to the original DPO baseline. By maintaining this increased diversity, \sys is better positioned to maximize the performance gains from TTS.
\section{Discussions and Limitations} \label{sec:discussion}

We discuss limitations and future directions of \sys.
Our TTS experiments are limited to $N=16$ parallel rollouts due to budget constraints. As our results in \autoref{fig:tts_curve} suggest that performance gains scale with $N$, exploring this behavior with a larger number of rollouts could more conclusively demonstrate the benefits of diversity preservation. Additionally, our end-to-end rollout strategy could be enhanced by incorporating more sophisticated search techniques like Tree-of-Thought~\citep{yao2023tree} or solution merging methods~\citep{luong2025advanced} to explore the solution space more effectively.

Third, due to computational constraints, the current \sys implementation adopts an offline preference learning setting. However, the proposed entropy-regularized preference learning framework naturally extends to online iterative settings, as in ~\citet{xiong2023iterative, guo2024direct}. Future work could investigate online versions of \sys-DPO and \sys-KTO, where the explicit diversity incentive may be crucially important for preventing mode collapse during repeated rounds of data collection and policy updates.

Finally, our framework extends naturally to scientific discovery domains such as molecular design~\citep{hou2025novo} and 3D structure prediction~\citep{jiao20243d}, which require generating diverse portfolios of candidates. In these scenarios, \sys can help navigate vast search spaces to uncover novel solutions, preventing the convergence to local optima common in standard preference learning.

\section{Conclusion} \label{sec:conclusion}

In this work, we introduce \sys, an entropy-enhanced preference optimization framework for multi-turn, tool-using agents on complex SWE tasks. By explicitly regularizing the preference objective to preserve policy diversity, \sys overcomes the diversity collapse issue. \sys achieves state-of-the-art results among open-weight models on the \swebench leaderboard, establishing a robust and effective method for building more powerful and reliable coding agents. We hope our work encourages further exploration into diversity-preserving alignment for building more capable and robust LLM agents.
\clearpage
\section*{Impact Statement}
This paper presents \sys, a framework designed to enhance the capabilities of open-weight Large Language Models (LLMs) in complex software engineering tasks. By improving the performance of publicly accessible models, we aim to foster transparency and democratize access to powerful coding assistants, reducing the performance gap between open-weight and proprietary closed-source models.

While automated coding agents offer significant productivity benefits, we acknowledge the potential for misuse in generating malicious code or introducing subtle vulnerabilities. However, our methodology focuses on correctness and alignment with verified test cases, which contributes to software reliability rather than degradation. Furthermore, all agent interactions during our experiments were conducted in strictly sandboxed environments to ensure safety.

From an environmental perspective, our approach utilizes offline preference optimization, which is computationally more efficient than online reinforcement learning methods that require extensive on-policy sampling. The datasets used in this work (\ie \swebench) are derived from public repositories; we are not aware of any personally identifiable information or offensive content within them.

\clearpage


\bibliography{main}
\bibliographystyle{icml2026}

\newpage
\appendix
\onecolumn

\section{Proof of \sref{Lemma}{lemma:reward}} \label{app:proof_lemma1}

We begin by expanding the sum of the regularized log-likelihood ratios:
\begin{align}
&\sum_{h=1}^{H} \zeta \log \frac{\pi^*(a_h|s_h)}{\pi_{\text{ref}}(a_h|s_h)^{\gamma/\zeta}} \nonumber \\
=& \sum_{h=1}^{H} (Q^*_{h}(s_h, a_h) - V^*_{h}(s_h) - \gamma \log \pi_{\text{ref}}(a_h|s_h)) \nonumber \\
=& \sum_{h=1}^{H} \bigg[ r(s_h, a_h) + \gamma \log \pi_{\text{ref}}(a_h|s_h)  + \mathbb{E}_{s' \sim \mathbb{P}_h(\cdot|s_h, a_h)}[V^*_{h+1}(s')] - V^*_{h}(s_h) - \gamma \log \pi_{\text{ref}}(a_h|s_h) \bigg] \nonumber \\
=& \sum_{h=1}^{H} r(s_h, a_h) - V^*_{1}(s_1) \nonumber  + \sum_{h=1}^{H-1} \left[ \mathbb{E}_{s' \sim \mathbb{P}_h(\cdot|s_h, a_h)}[V^*_{h+1}(s')] - V^*_{h+1}(s_{h+1}) \right],
\end{align}

where the last equation using the fact $V_{H+1}(s) = 0$. Given the deterministic nature of our software engineering task, the expectation term collapses:
$
\sum_{h=1}^{H-1} [ \mathbb{E}_{s' \sim p(\cdot|s_h, a_h)}[V^*_{h+1}(s')] - V^*_{h+1}(s_{h+1}) ] = 0.
$

Rearranging the terms yields the final relationship stated in the lemma.

\section{\sys-KTO loss}
\label{app:kto_loss}
Alternatively, the policy can be learned by optimizing a KTO-style objective. Consider a dataset $\mathcal{D}$ where each trajectory $\tau$ is labeled as either ``desirable'' ($y=+$) or ``undesirable'' ($y=-$). We define the implicit reward for a response $y$ given context $x$ as:
\begin{equation}
    r_\theta(x, y) = \sum_{h=1}^{H} \log \frac{\pi_{\theta}(a_h|s_h)}{\pi_{\text{ref}}(a_h|s_h)^{\gamma/\zeta}}
\end{equation}
The \text{\sys-KTO} loss is designed to maximize the margin between desirable and undesirable trajectories relative to a reference point $z_0$. The objective is defined as:
\begin{equation}
    \mathcal{L}_{\text{\sys-KTO}}(\theta) = \mathbb{E}_{(x, y) \sim \mathcal{D}} \left[ \lambda_y - v(x, y) \right]
\end{equation}
where the value function $v(x, y)$ is given by:
\begin{equation}
    v(x, y) = 
    \begin{cases} 
        \lambda_+ \sigma\Big(\zeta \big(r_\theta(x, y) - z_0 \big)\Big) & \text{if } y=+, \\
        \lambda_- \sigma\Big(\zeta \big(z_0 - r_\theta(x, y) \big)\Big) & \text{if } y=-.
    \end{cases}
\end{equation}
Here, $\lambda_+$ and $\lambda_-$ are hyperparameters that weight the loss for desirable and undesirable examples, respectively. The reference point $z_0$ is defined as:
\begin{equation}
    z_0 = \mathbb{E}_{x\sim \mathcal{D}, \tau \sim \pi(\cdot|x)} \Bigg[ \sum_{h=1}^{H} \bigg( -H(\pi(\cdot|s_h)) 
    + \frac{\gamma}{\zeta}H(\pi(\cdot|s_h), \pi_{\text{ref}}(\cdot|s_h)) \bigg) \Bigg].
\end{equation}

While $z_0$ theoretically depends on the evolving policy $\pi$, we treat it as a detached constant during the gradient computation step (\ie we do not backpropagate through the expectation over $\pi$) following \citet{ethayarajh2024kto}. 


\section{Diversity Analysis}

\subsection{Equivalence of Conditional Probability}
\label{app:proof_bandit}

A response $y$ structurally consists of the model's generated tokens $y'=(y'_1, \dots, y'_H)$ (code) and the resulting observation $o=(o_1, \dots, o_{H-1})$ (execution feedback).
By the chain rule, the probability of the trajectory is $P(y|x) = \tilde{\pi}(y'|x) P(o|x, y')= \tilde{\pi}(y'_1|x)P(o_1|x, y'_1) \dots \tilde{\pi}(y'_H|x, y'_1, o_1, \dots, y'_{H-1}, o_{H-1})$. In the context of software engineering tasks, the environment exhibits \textit{deterministic transition dynamics}. Thus, $P(o|x, y')$ is equal to 1. Optimizing the trajectory probability $P(y|x)$ is therefore mathematically equivalent to optimizing the generative policy $ \tilde{\pi}(y'|x)$.

\subsection{Derivation of the Optimal Policy}
\label{app:proof_optimal}
Through policy re-parametrization, we seek to maximize the following objective:
\begin{equation}
\mathcal{J}(\tilde{\pi}) = \sum_y \tilde{\pi}(y|x) r(x,y) - \zeta \sum_y \tilde{\pi}(y|x) \log \tilde{\pi}(y|x) + \gamma \sum_y \tilde{\pi}(y|x) \log  \tilde{\pi}_{\mathrm{ref}}(y|x).
\end{equation}
We construct the Lagrangian with a constraint $\sum_y \tilde{\pi}(y|x) = 1$:
\begin{equation}
\mathcal{L}(\tilde{\pi}, \eta) = \mathcal{J}(\tilde{\pi}) + \eta \left(1 - \sum_y \tilde{\pi}(y|x) \right).
\end{equation}
Taking the derivative w.r.t $\tilde{\pi}(y|x)$ and setting to 0:
\begin{equation}
\frac{\partial \mathcal{L}}{\partial \tilde{\pi}(y|x)} = r(x,y) - \zeta(1 + \log \tilde{\pi}(y|x)) + \gamma \log  \tilde{\pi}_{\mathrm{ref}}(y|x) - \eta = 0.
\end{equation}
Solving for $\tilde{\pi}(y|x)$:
\begin{equation}
\log \tilde{\pi}(y|x) = \frac{1}{\zeta} \left( r(x,y) + \gamma \log  \tilde{\pi}_{\mathrm{ref}}(y|x) - \eta - \zeta \right).
\end{equation}
Exponentiating and absorbing constants into a partition function $Z(x)$ yields the result:
\begin{equation}
\tilde{\pi}^*(y|x) = \frac{1}{Z(x)}  \tilde{\pi}_{\mathrm{ref}}(y|x)^{\frac{\gamma}{\zeta}} \exp\left(\frac{r(x,y)}{\zeta}\right). 
\label{eqn:optimal_policy}
\end{equation}

Formally, the optimization above is a \emph{constrained} problem: because the environment has deterministic (fixed) transition dynamics, only trajectories consistent with those dynamics are feasible. Consequently, for a given state (or context) $x$, the response/trajectory $y$ cannot range over the entire trajectory space; instead, $\tilde{\pi}(y\mid x)$ must assign zero probability to any $y$ that violates the environment transition constraint.

Crucially, both $\tilde{\pi}$ and the reference policy $\tilde{\pi}_{\mathrm{ref}}$ are learned in the \emph{same} environment, and therefore share the \emph{same support} over valid trajectories (\ie $\tilde{\pi}_{\mathrm{ref}}(y\mid x)=0$ whenever $y$ is infeasible). Since $\tilde{\pi}_{\mathrm{ref}}(y\mid x)$ enters multiplicatively in \sref{Eqn.}{eqn:optimal_policy}, it implicitly restricts the solution to the feasible set. Hence, the derived optimal policy $\tilde{\pi}^*$ remains unchanged in form and is valid under the environment-induced constraint.

\subsection{Proof of \sref{Proposition}{proposition:diversity}}
\label{app:proof_claim}
We now analyze how the hyperparameters $\gamma$ and $\zeta$ influence the diversity of the generated responses.

Specifically, we consider the log-ratio of the optimal policy $\tilde{\pi}^*$ to the reference policy $\pi_{\mathrm{ref}}$. Using the result from \sref{Lemma}{lemma:optimal_policy}:
\begin{align}
\log \frac{\tilde{\pi}^*(y|x)}{ \tilde{\pi}_{\mathrm{ref}}(y|x)} &= \log \left( \frac{ \tilde{\pi}_{\mathrm{ref}}(y|x)^{\gamma/\zeta} \exp(r(x,y)/\zeta)}{Z(x)  \tilde{\pi}_{\mathrm{ref}}(y|x)} \right) \\
&= \frac{r(x,y)}{\zeta} + \left(\frac{\gamma}{\zeta} - 1\right) \log  \tilde{\pi}_{\mathrm{ref}}(y|x) - \log Z(x).
\label{eqn:update_rule}
\end{align}
This equation describes the ``update'' or ``boost'' applied to the reference policy to obtain the optimal policy.

\paragraph{Comparison with Standard RLHF/DPO.}
Standard DPO corresponds to the case where the KL-divergence penalty is used alone, which implies $\gamma = \zeta$ (\ie $\gamma/\zeta = 1$). In this case, the update rule simplifies to:
\begin{equation}
\log \frac{\tilde{\pi}^*_{\text{DPO}}(y|x)}{ \tilde{\pi}_{\mathrm{ref}}(y|x)} = \frac{r(x,y)}{\zeta} - \log Z_{\text{DPO}}(x).
\end{equation}
Here, the relative boost depends \emph{only} on the reward $r(x,y)$. If two responses $y_1$ and $y_2$ have equal reward $r(x,y_1) = r(x,y_2)$ (\ie In RLVR, the correct answer usually receives the same reward), they receive the exact same log-probability boost, regardless of their initial likelihood under $ \tilde{\pi}_{\mathrm{ref}}$.

\paragraph{\sys Regime ($\gamma < \zeta$).}
In \sys, we configure $\gamma < \zeta$, such that the coefficient $(\frac{\gamma}{\zeta} - 1)$ is strictly negative. Let us compare two responses, $y_{\text{common}}$ and $y_{\text{rare}}$, which are both ``correct'' (equal high reward $r$) but have different probabilities under the reference model:
\begin{equation}
\tilde{\pi}_{\mathrm{ref}}(y_{\text{common}}|x) > \tilde{\pi}_{\mathrm{ref}}(y_{\text{rare}}|x) \implies \log \tilde{\pi}_{\mathrm{ref}}(y_{\text{common}}|x) > \log \tilde{\pi}_{\mathrm{ref}}(y_{\text{rare}}|x).
\end{equation}
Since $\log  \tilde{\pi}_{\mathrm{ref}}$ is negative and $(\frac{\gamma}{\zeta} - 1)$ is negative, the term $(\frac{\gamma}{\zeta} - 1) \log  \tilde{\pi}_{\mathrm{ref}}(y|x)$ is positive and monotonically decreasing with respect to $ \tilde{\pi}_{\mathrm{ref}}(y|x)$.

Calculating the difference in the update magnitude $\Delta(y) = \log \tilde{\pi}^*(y|x) - \log  \tilde{\pi}_{\mathrm{ref}}(y|x)$ for the two responses:
\begin{align}
\Delta(y_{\text{rare}}) - \Delta(y_{\text{common}}) &= \left(\frac{\gamma}{\zeta} - 1\right) \left[ \log  \tilde{\pi}_{\mathrm{ref}}(y_{\text{rare}}|x) - \log  \tilde{\pi}_{\mathrm{ref}}(y_{\text{common}}|x) \right] > 0.
\end{align}
The inequality holds because we are multiplying a negative coefficient by a negative difference. This implies $\Delta(y_{\text{rare}}) > \Delta(y_{\text{common}})$.

Therefore, under \sys, the rare correct response receives a \emph{larger} relative boost than the common correct response. 
This mechanism counteracts the ``rich-get-richer'' dynamic of standard RLHF, thereby preserving and promoting diversity in the distribution of correct solutions.

\section{Experiment Details} \label{app:implementation}

In this section, we provide additional details about the experiment details of \sys.

\subsection{Dataset Details}
\textbf{Training Data.} For SFT, we use the SWE-Smith dataset~\citep{yang2025swe}, selecting only Python-based instances with problem statements that align with the \swebench format. For preference learning, we employ the R2E-Gym-subset~\citep{jain2025r2e}, which we selected because it contains no repository overlap with our test sets, thereby preventing data leakage.

\textbf{Evaluation Data.} We evaluate \sys on the complete official test sets for both \verified and \lite. Detailed statistics for all datasets are provided in~\autoref{tab:dataset-details}.

\begin{table}[h!]
    \centering
    \caption{
        \textbf{Dataset Statistics}: The table presents the number of training and test samples for each dataset, along with the source of the dataset.
    }
    \label{tab:dataset-details}
    \resizebox{1.0\columnwidth}{!}{ 
    \begin{tabular}{l c c l}
    \toprule
    \textbf{Dataset} & \textbf{\# Train Samples} & \textbf{\# Test Samples} & \textbf{Source} \\
    \midrule
    SWE-Smith                 & 8736 &  -         & \url{https://huggingface.co/datasets/r2e-edits/swesmith-clean} \\
    R2E-Gym-subset            & 4578 &  -          & \url{https://huggingface.co/datasets/R2E-Gym/R2E-Gym-Subset} \\
    SWE-bench Verified        & - & 500         & \url{https://huggingface.co/datasets/princeton-nlp/SWE-bench_Verified} \\
    SWE-bench Lite            & - & 300         & \url{https://huggingface.co/datasets/princeton-nlp/SWE-bench_Lite} \\
    \bottomrule
    \end{tabular}
    }
\end{table}

\subsection{Scaffold Details}

In our experiments, we utilize the standard R2E scaffold, which is recognized for its flexibility, ease of use, and robust performance. This scaffold equips the agent with four essential tools: \texttt{file\_editor} for file editing, \texttt{execute\_bash} for running bash commands, \texttt{search} for file and code retrieval, and \texttt{finish} to conclude the task. The complete system prompt, detailing the functionality and parameters of each tool, is provided below.

\begin{tcolorbox}[colback=blue!5!white, colframe=blue!75!black, title=System Prompt of the Scaffold, fonttitle=\bfseries, breakable]
    You are a programming agent who is provided a GitHub issue and repository bash environment and is tasked to solve certain tasks (e.g., file localization, testcase generation, code repair, and editing, etc) to resolve the issue.

  We have access to the following functions:

  --- BEGIN FUNCTION \#1: \texttt{file\_editor} ---
  
  \textbf{Description}: Custom editing tool for viewing, creating, and editing files.
    \begin{itemize}[leftmargin=*]
    \item State is persistent across command calls and discussions with the user
    \item If path is a file, view displays the result of applying \texttt{cat -n}. If path is a directory, view lists of non-hidden files and directories up to 2 levels deep
    \item The create command cannot be used if the specified path already exists as a file
    \item If a command generates a long output, it will be truncated and marked with \texttt{<response clipped>}
    \item The \texttt{undo\_edit} command will revert the last edit made to the file at path
    \end{itemize}

  \textbf{Notes for using the \texttt{str\_replace} command}:
    \begin{itemize}[leftmargin=*]
    \item The \texttt{old\_str} parameter should match EXACTLY one or more consecutive lines from the original file. Be mindful of whitespaces!
    \item If the \texttt{old\_str} parameter is not unique in the file, the replacement will not be performed. Make sure to include enough context in \texttt{old\_str} to make it unique
    \item The \texttt{new\_str} parameter should contain the edited lines that should replace the \texttt{old\_str}
    \end{itemize}

  \textbf{Parameters}:
    \begin{enumerate}[leftmargin=*]
    \item \texttt{command} (string, required) \\
  Allowed values: \texttt{[view, create, str\_replace, insert, undo\_edit]}. The command to run.
    \item \texttt{path} (string, required) \\
  Absolute path to file or directory, e.g. \texttt{/testbed/file.py} or \texttt{/testbed}.
    \item \texttt{file\_text} (string, optional) \\
  Required for the \texttt{create} command. Contains the content of the file to be created.
    \item \texttt{old\_str} (string, optional) \\
  Required for the \texttt{str\_replace} command. The exact string in the path to replace.
    \item \texttt{new\_str} (string, optional)
        \begin{itemize}[leftmargin=*]
        \item Optional for the \texttt{str\_replace} command to specify the replacement string.
        \item Required for the \texttt{insert} command to specify the string to insert.
        \end{itemize}
    \item \texttt{insert\_line} (integer, optional) \\
  Required for the \texttt{insert} command. The \texttt{new\_str} will be inserted after the line number specified here.
    \item \texttt{view\_range} (array, optional)
        \begin{itemize}[leftmargin=*]
        \item Optional for the \texttt{view} command (when path is a file).
        \item If provided, specifies the line range to view, e.g. \texttt{[11, 12]} shows lines 11 and 12.
        \item \texttt{[start\_line, -1]} will show all lines from \texttt{start\_line} to the end of file.
        \end{itemize}
    \item \texttt{concise} (boolean, optional)
        \begin{itemize}[leftmargin=*]
        \item Optional for the \texttt{view} command.
        \item Displays a concise skeletal view of the file. If set to \texttt{False}, it displays the full content in the specified \texttt{view\_range}.
        \end{itemize}
    \end{enumerate}

  --- END FUNCTION \#1 ---

  --- BEGIN FUNCTION \#2: \texttt{execute\_bash} ---
  
  \textbf{Description}: Execute a bash command in the terminal.

  \textbf{Behavior notes}:
    \begin{itemize}[leftmargin=*]
    \item If a command may run indefinitely (long-running), consider running it in the background and redirecting output, e.g. \verb|python3 app.py > server.log 2>&1 &|.
    \item If the bash command returns exit code -1, it means the process is still running. The assistant may:
        \begin{itemize}[leftmargin=*]
        \item Call this function again with the command as an empty string ("") to retrieve additional logs.
        \item Send more input to STDIN of the running process by calling this function again with the command set to the text input.
        \item Send \texttt{command="ctrl+c"} to interrupt the currently running process.
        \end{itemize}
    \item If the command times out, it will be interrupted (SIGINT). The assistant may then retry or do further steps if needed.
    \end{itemize}

  \textbf{Parameters}:
    \begin{enumerate}[leftmargin=*]
    \item \texttt{cmd} (string, required) \\
  The bash command (and optional arguments) to execute.
        \begin{itemize}
        \item Can be empty ("") to retrieve more logs if the process is still running.
        \item Can be "ctrl+c" to interrupt the running process.
        \end{itemize}
    \end{enumerate}

  --- END FUNCTION \#2 ---

  --- BEGIN FUNCTION \#3: \texttt{search} ---
  
  \textbf{Description}: Search for a term in a directory or a single file.
    \begin{itemize}[leftmargin=*]
    \item If path is a directory (or unspecified, default is \texttt{.} ), it recursively searches all non-hidden files and directories for the search term.
    \item If path points to a file, it runs a \texttt{grep -n} in that file to show line numbers matching the search term.
    \item If more than 100 files match in a directory search, results are truncated, and the tool will inform you to narrow your search.
    \item If no matches are found, it will inform you as well.
    \end{itemize}

  \textbf{Parameters}:
    \begin{enumerate}[leftmargin=*]
    \item \texttt{search\_term} (string, required) \\
  The term or string to search for in files.
    \item \texttt{path} (string, optional) \\
  The file or directory to search in. Defaults to \texttt{.} if not specified.
    \end{enumerate}

  --- END FUNCTION \#3 ---

  --- BEGIN FUNCTION \#4: \texttt{finish} ---
  
  \textbf{Description}: Finish the interaction once the task is complete or if no further progress can be made.

  \textbf{Behavior notes}:
    \begin{itemize}[leftmargin=*]
    \item The submit command finalizes your output.
    \end{itemize}

  \textbf{Parameters}:
    \begin{enumerate}[leftmargin=*]
    \item \texttt{command} (string, required) \\
  Currently allowed value: \texttt{[submit]}.
    \item \texttt{result} (string, optional) \\
  The result text or final message to submit. Defaults to an empty string if not provided.
    \end{enumerate}

  --- END FUNCTION \#4 ---

  If you choose to call a function ONLY reply in the following format with NO suffix:

\begin{verbatim}
<function=example_function_name>
<parameter=example_parameter_1>value_1</parameter>
<parameter=example_parameter_2>
This is the value for the second parameter
that can span
multiple lines
</parameter>
</function>
\end{verbatim}

  \texttt{<IMPORTANT>}
  Reminder:
  \begin{itemize}[leftmargin=*]
  \item Function calls MUST follow the specified format, start with \texttt{<function=} and end with \texttt{</function>}
  \item Required parameters MUST be specified
  \item Only call one function at a time
  \item VERY IMPORTANT: Each response must include both reasoning (as natural text) and function call (in the above format) to solve the task.
  \end{itemize}
\end{tcolorbox}

\begin{figure}[ht]
  \centering
  \includegraphics[width=0.8\textwidth]{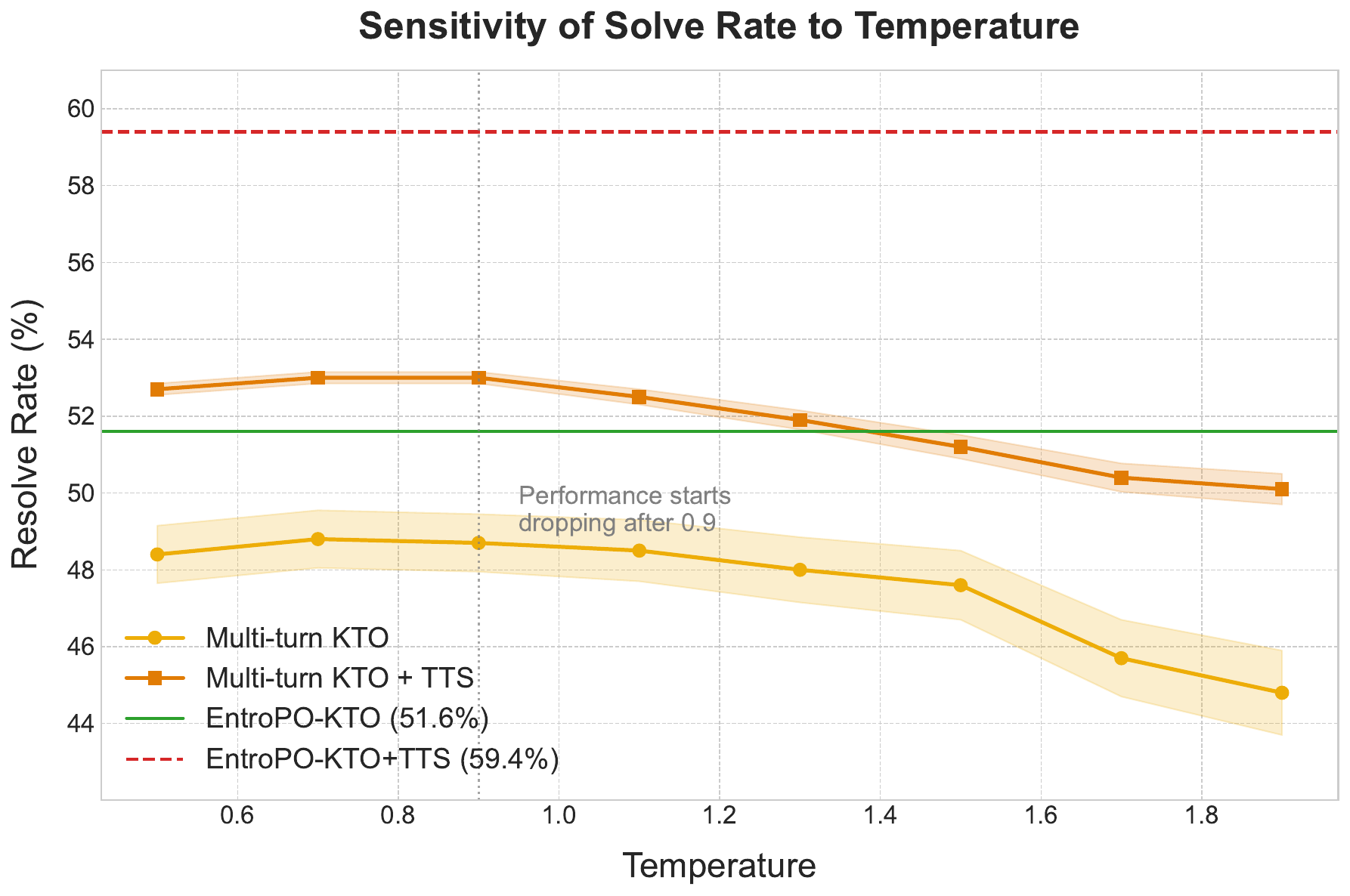}
  \caption{\textbf{Impact of Temperature on Performance.} Performance of multi-turn KTO and KTO+TTS on \verified with varying temperature, compared to \sys with a fixed temperature of 0.7. Increasing temperature fails to match the performance of \sys and degrades performance past 0.9.}
  \label{fig:temperature_sensitivity}
\end{figure}

\subsection{Running Environment}
Our implementation relies on the LLaMA-Factory~\citep{zheng2024llamafactory} for model training and SGLang~\citep{zheng2024sglang} for efficient inference deployment. All experiments are performed on a server configured with four 32-core AMD EPYC 7702 CPUs, 8 NVIDIA H100 (80GB) GPUs, and 4 NVIDIA A100 (40GB) GPUs. The H100 GPUs are dedicated to the primary training and inference workloads, while the A100 GPUs provide supplementary computational support during inference.

\subsection{Training and Inference Details} \label{app:training_inference_details}
\textbf{Training Pipeline.} Our training process proceeds in two stages. The first stage is SFT, where we teach the base model to use tools reliably. We generate a dataset $\mathcal{D}_{\mathrm{SFT}}$ of successful interaction trajectories using a strong teacher model, \ie GLM-4.5. The student model is then fine-tuned on these examples to learn stable tool-use patterns of the scaffold. The second stage is preference learning with \sys. After SFT, we generate a new pool of trajectories by rolling out both the SFT-tuned student model and teacher model. We use a SWE dataset with commit-corresponding test cases for each instance so that we can get the preference label for each trajectory. If the final patch passes the test cases, it is labeled as preferred. Otherwise, it is labeled as not preferred. From this pool, we create a preference dataset $\mathcal{D}_{\mathrm{pref}}$ by pairing trajectories for the same problem, labeling the one with the higher score as preferred. The SFT model is then further fine-tuned on this dataset using our entropy-enhanced objective.

\textbf{Training Hyperparameters.} We configure our training process as follows. For LoRA, we set \texttt{lora\_rank=8}, \texttt{lora\_alpha=16}, and apply it to all available modules (\texttt{lora\_target=all}). For the GLM-4.5-Air model, we use 4-bit QLoRA quantization. Across all models, we use a context length of 18,000 tokens, a learning rate of 1e-5, and a warmup ratio of 0.1. 
\revise{This context length is chosen to avoid Out of Memory errors during training as a longer context length would require much more memory.}
The batch size is set to 4 for GLM-4.5-Air and 16 for all other models. Both the SFT and preference optimization stages are trained for a single epoch using these settings.

\textbf{Inference Parameters.} During inference, each rollout is permitted a maximum of 200 environment interaction steps and a total generation length of up to 131,072 tokens. We use a sampling temperature of 0.7, \texttt{top\_k} of 20, and \texttt{top\_p} of 0.8 for all models, which is the recommended setting from Qwen3 model card.

\textbf{Hybrid Trajectory Selection.} Our hybrid selector employs a multi-stage filtering process to identify the best trajectory from the generated candidates $\{\tau^{(n)}\}_{n=1}^N$. The verifier $p_\phi(x,\tau)\in[0,1]$ is a learned scorer trained on $\mathcal{D}_{\mathrm{pref}}$ with supervised learning. We select the Qwen3-Coder-30B as our verifier model.
Unlike prior work~\citep{jain2025r2e} that uses $p_\phi$ as the major ranking criterion, we use it as a conservative filter. The process is as follows:
\begin{enumerate}
    \item \textbf{Finished score (model-free)}: We first discard any trajectories that are truncated due to exceeding the step or token limits: keep only $\mathbbold{1}(\mathrm{finished}(\tau)=1)$.
    \item \textbf{Regression test score (model-free)}: Next, we filter out trajectories that fail the regression tests generated by the R2E-Gym framework: $\mathbbold{1}(\mathrm{regress\_free}(\tau)=1)$.
    \item \textbf{Verifier probability (model-based)}: We then apply a verifier model and remove trajectories with a probability below a threshold of $\eta=0.01$. We choose a conservative threshold because we empirically observe that many valid solutions do not receive high probability scores, but scores below 0.01 are a strong indicator of a flawed trajectory. We keep $\{\tau\in\mathcal{S}: p_\phi(x,\tau)\ge \eta\}$.
\end{enumerate}
If any filtering step results in an empty set of candidates, we revert to the candidate pool from the previous step.

After filtering, we apply a final model-free heuristic for selection. We select the trajectory with the \textbf{most} environment interaction steps for \verified and the \textbf{fewest} for \lite. This distinction is motivated by the nature of the benchmarks and our observation. 
\revise{
  For \verified, every instance has been manually verified by human engineers at OpenAI~\citep{chowdhury2024swebenchverified} to have accurate problem statements and robust environments. Thus, a longer successful trajectory can be a positive signal, potentially indicating more comprehensive reasoning, the addition of more thorough test cases, or the consideration of more corner cases before submitting a final patch.
  However, for \lite, the situation is different. It is known to contain instances with misleading or incomplete problem statements. \citet{agentless} and \citet{chowdhury2024swebenchverified} note that the original SWE-bench dataset contains underspecified problem statements and problematic environment setups that cause some unit tests to fail regardless of the solution. In such cases, a long trajectory can signal that the agent is misled by a vague prompt or is ``hallucinating'' complexity in response to incorrect unit test feedback. Therefore, we adopt the strategy from prior work~\citep{agarwal2025first,hassid2025don} to prefer shorter solutions, which helps mitigate the negative impact of these problematic instances. We conduct ablation studies on the impact of this strategy in \autoref{app:additional_experiments} which shows that this strategy is effective on \lite.
}

\subsection{Additional Experiments} \label{app:additional_experiments}
\textbf{Impact of Temperature on Performance.}
To investigate whether increased sampling diversity can replicate the benefits of our entropy-regularized approach, we evaluate the multi-turn KTO and multi-turn KTO+TTS models under varying temperatures. We test the Qwen3-Coder-30B model on \verified with temperatures ranging from 0.5 to 1.8 and compare its performance to \sys-KTO and \sys-KTO+TTS, which use a fixed temperature of 0.7. As shown in \autoref{fig:temperature_sensitivity}, increasing the temperature provides no performance benefit. In fact, performance begins to decline beyond a temperature of 0.9, as excessive sampling randomness undermines the precision required for SWE tasks involving tool use and code editing. These results demonstrate that merely increasing temperature is not a substitute for principled entropy regularization, as it fails to match the performance gains achieved by \sys.

\begin{figure*}[t]
  \centering
  \includegraphics[width=0.48\textwidth]{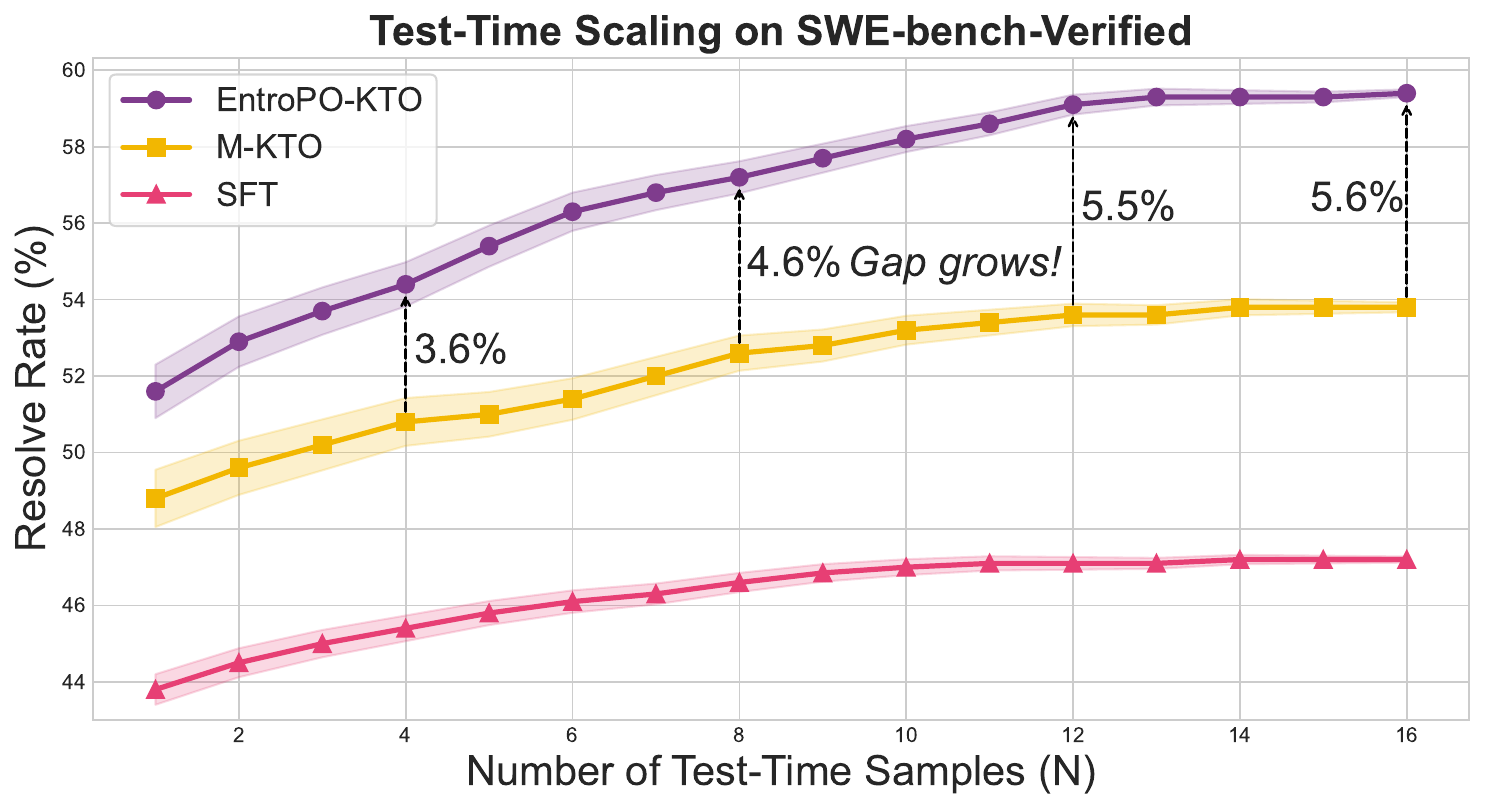}
  \includegraphics[width=0.48\textwidth]{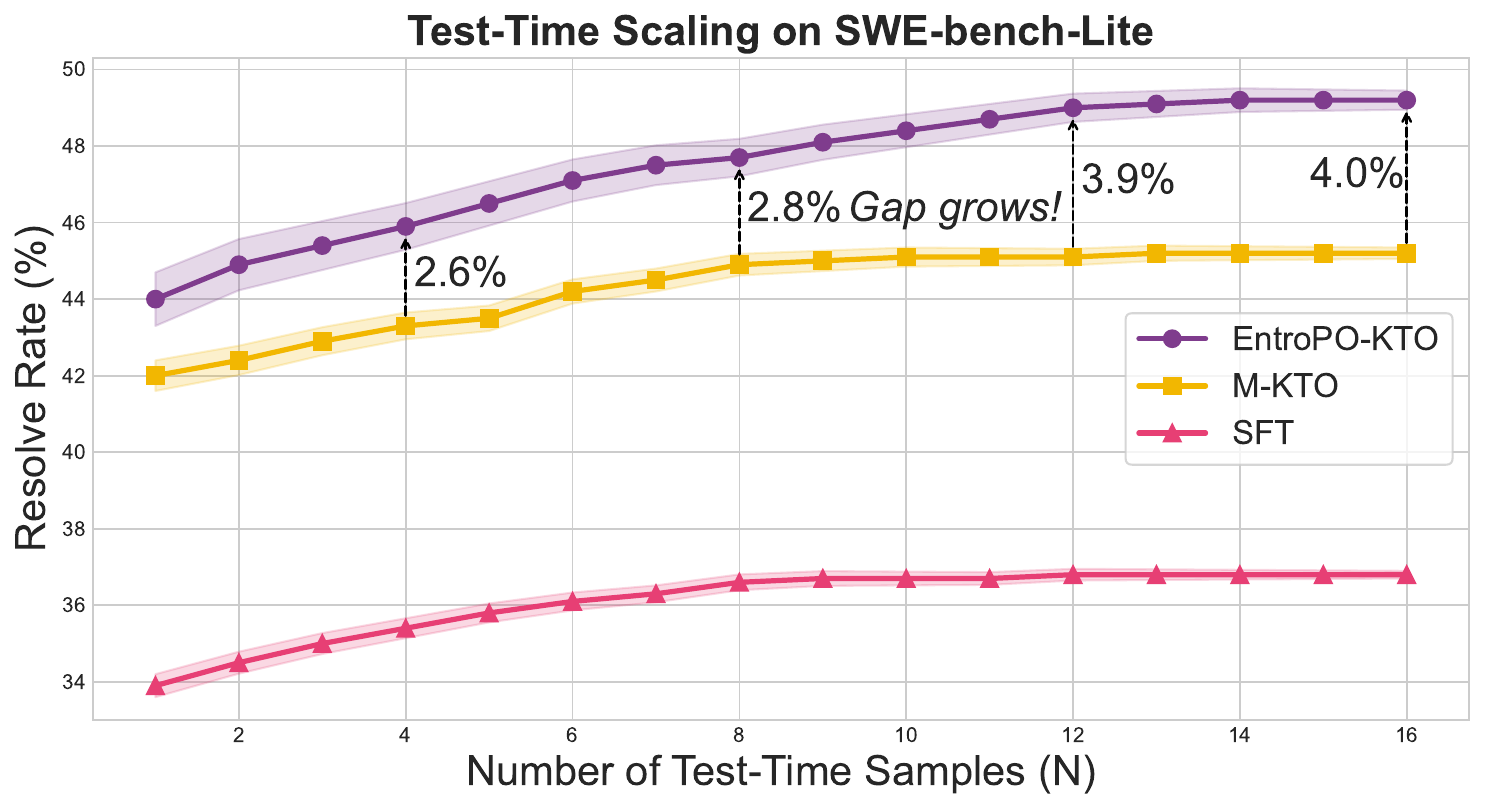}
  \caption{\textbf{The Impact of Entropy Regularization on Test-Time Scaling.} Performance of \sys-KTO, M-KTO, and SFT on \verified (left) and \lite (right) as the number of parallel rollouts ($N$) increases. \sys's entropy regularization consistently yields better scaling.}
  \label{fig:tts_curve_dpo}
\end{figure*}

\revise{
\textbf{Impact of Entropy Regularization on KTO.}
To investigate whether our entropy-regularized approach can benefit KTO, we conduct experiments on the Qwen3-Coder-30B model with the \sys-KTO and M-KTO~\citep{xiong2025building} models. We test the Qwen3-Coder-30B model on \verified and \lite with different number of parallel rollouts ($N$) and compare its performance to \sys-KTO, M-KTO, and SFT. As shown in \autoref{fig:tts_curve_dpo}, \sys-KTO consistently outperforms M-KTO and SFT. When $N=1$, \sys-KTO can outperform both SFT and M-KTO, showing its advantage can be independent of TTS. As $N$ increases, the performance gap between \sys-KTO and M-KTO widens, similar to our findings in \autoref{fig:tts_curve}.
}

\revise{
\textbf{Impact of Step Heuristic on \lite.}
We adopt different step heuristics for \sys-KTO with model Qwen3-Coder-30B on \lite as follows: 1) Prioritize the trajectory with the most environment interaction steps. 2) Prioritize the trajectory with the fewest environment interaction steps. 3) Randomly select a trajectory as the step heuristic. The results are shown in \autoref{tab:step_heuristic_lite}. We can observe that when using the fewest steps heuristic, the performance of \sys-KTO is best, which verifies our motivation in \autoref{app:training_inference_details}. However, even using the longest steps heuristic or random selection heuristic, the performance of \sys-KTO is still better than the best open-source submission on \lite (CodeFuse-CGM with 44.0\% resolve rate), which demonstrates the robustness of our overall framework.

\textbf{Heuristics for a Random Real-World SWE Problem.}
  The above analysis leads to a practical guide for choosing a heuristic for a random real-world SWE problem:
  \begin{itemize}
      \item \textbf{If the developer trust the specification of the SWE problem:} Favor Longest Steps.
      \item \textbf{If the specification is vague/noisy:} Favor Shortest Steps.
      \item \textbf{If the developer has no idea (The Default Strategy):} We recommend favoring Longest Steps.
  \end{itemize}
  As shown in \autoref{tab:step_heuristic_lite}, when we applied the ``Longest Steps'' heuristic to \lite (where it is empirically suboptimal), the performance (47.3\%) was comparable to Random Selection (47.1\%). This means that even if the heuristic is ``suboptimal'' for the data distribution, it causes no significant harm. However, on high-quality data (\verified), using Longest Steps yields significant gains. Therefore, prioritizing the longest trajectory can be a default choice---it captures the upside on good data without degrading performance on noisy data.

\begin{table}[h!]
    \centering
    \caption{
        \textbf{Impact of Step Heuristic on \lite.}: The table presents the performance of different step heuristics on \lite. We compare prioritizing trajectories with the most steps, fewest steps, and random selection.
    }
    \label{tab:step_heuristic_lite}
    \resizebox{0.5\columnwidth}{!}{
    \begin{tabular}{l c c c}
    \toprule
    & \textbf{Most Steps} & \textbf{Fewest Steps} & \textbf{Random} \\
    \midrule
    \textbf{Resolve Rate (\%)} & 47.3 ($\pm$ 0.3) & 49.2 ($\pm$ 0.7) & 47.1 ($\pm$ 0.6) \\
    \bottomrule
    \end{tabular}
    }
\end{table}

}

\textbf{Diversity Analysis.} \label{app:diversity_analysis}
Following \citet{hongcuriosity}, we evaluate trajectory diversity by measuring cosine similarity across multiple runs of the same problem.
We use the Qwen3-4B-Instruct-2507 model for this analysis due to its efficiency and long context support (262k tokens).
For each problem instance, we sample 10 independent runs, from which we construct all pairwise trajectory combinations (yielding 45 pairs). Each trajectory is represented by the hidden representation of the final token from the model’s last layer. We compute the cosine similarity for each pair and average these values to obtain a per-problem similarity score. Finally, we report the average of these scores across all problems in the \verified dataset.
The results are presented in \autoref{tab:diversity_analysis}.
We observe that \sys-DPO and \sys-KTO exhibit lower average cosine similarities (0.8635 and 0.8724) compared to their standard counterparts, M-DPO (0.9125) and M-KTO (0.9053).
These lower similarity scores indicate a higher degree of diversity in the trajectories generated by \sys, protecting against the mode collapse often observed in preference learning and enabling more effective test-time scaling.

\begin{table}[htbp]
    \centering
    \caption{
        \textbf{Diversity Analysis on \verified}: We report the average cosine similarity of trajectory pairs for different methods. Lower similarity indicates higher diversity.
    }
    \label{tab:diversity_analysis}
    \resizebox{0.65\columnwidth}{!}{
    \begin{tabular}{l c c c c}
    \toprule
    \textbf{Method} & \textbf{M-DPO} & \textbf{\sys-DPO} & \textbf{M-KTO} & \textbf{\sys-KTO} \\
    \midrule
    \textbf{Cosine Similarity} & 0.9125 & 0.8635 & 0.9053 & 0.8724 \\
    \bottomrule
    \end{tabular}
    }
\end{table} 

\end{document}